%% file: main.tex
\documentclass[10pt,twocolumn,letterpaper]{article}

\usepackage{cvpr}              

\usepackage{graphicx}
\usepackage{amsmath}
\usepackage{amssymb}
\usepackage{booktabs}
\usepackage[pagebackref,breaklinks,colorlinks]{hyperref}
\usepackage[accsupp]{axessibility}

\usepackage[capitalize]{cleveref}
\crefname{section}{Sec.}{Secs.}
\Crefname{section}{Section}{Sections}
\Crefname{table}{Table}{Tables}
\crefname{table}{Tab.}{Tabs.}

\def\cvprPaperID{8338} 
\def\confName{CVPR}
\def\confYear{2022}

\begin{document}

\title{Modeling Indirect Illumination for Inverse Rendering}

\author{
Yuanqing Zhang$^{1,2}$
\quad
Jiaming Sun$^1$
\quad
Xingyi He$^1$
\quad
Huan Fu$^2$
\quad
Rongfei Jia$^2$
\quad
Xiaowei Zhou$^{1*}$
\quad
\\[1.5mm]
$^1$Zhejiang University
\quad
$^2$Tao Technology Department, Alibaba Group
}

\maketitle

\begin{abstract}
Recent advances in implicit neural representations and differentiable rendering make it possible to simultaneously recover the geometry and materials of an object from multi-view RGB images captured under unknown static illumination. Despite the promising results achieved, indirect illumination is rarely modeled in previous methods, as it requires expensive recursive path tracing which makes the inverse rendering computationally intractable. In this paper, we propose a novel approach to efficiently recovering spatially-varying indirect illumination. The key insight is that indirect illumination can be conveniently derived from the neural radiance field learned from input images instead of being estimated jointly with direct illumination and materials. By properly modeling the indirect illumination and visibility of direct illumination, interreflection- and shadow-free albedo can be recovered. The experiments on both synthetic and real data demonstrate the superior performance of our approach compared to previous work and its capability to synthesize realistic renderings under novel viewpoints and illumination. Our code and data are available at \href{https://zju3dv.github.io/invrender/}{https://zju3dv.github.io/invrender/}.

\end{abstract}

\vspace{-1em}
\let\thefootnote\relax\footnotetext{The authors from Zhejiang University are affiliated with the State Key Lab of CAD\&CG. This work was done when Yuanqing Zhang was an intern at Alibaba Group. $^*$Corresponding author: Xiaowei Zhou.}

\input{01_introduction.tex}
\input{02_related_work.tex}

\input{03_method.tex}
\input{04_experiment.tex}

\input{05_conclusion.tex}

\noindent\textbf{Acknowledgements:} 
This work was supported by NSFC (No.~62172364) and Alibaba Group through Alibaba Innovative Research Program.
\vspace{3em}

{\small
\bibliographystyle{ieee_fullname}
\bibliography{main}
}

\end{document}

%% file: 01_introduction.tex
\begin{figure}[t]
\centering
\setlength{\abovecaptionskip}{2.0mm}
\includegraphics[width=1\linewidth]{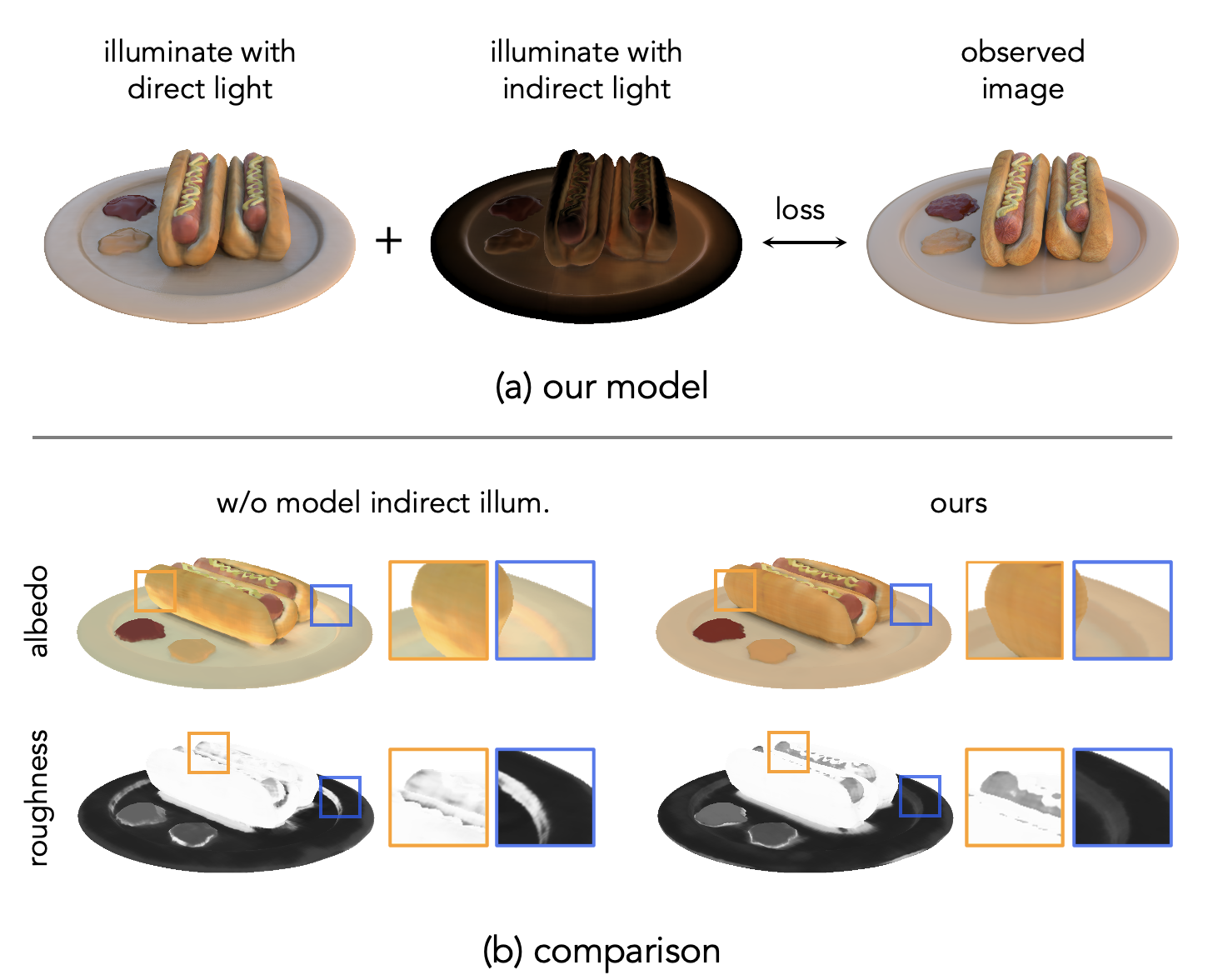}
\caption{To precisely recover SVBRDF (parameterized as albedo and roughness) from multi-view RGB images, we propose an efficient approach to reconstruct spatially varying indirect illumination and combine it with environmental light evaluated by visibility as the full light model (a). The example in (b) demonstrates that without modeling indirect illumination, its rendering effects are baked into the estimated albedo to compensate for the incomplete light model and also result in artifacts in the estimated roughness.
}
\label{fig:banner}
\end{figure}

\section{Introduction}
Recovering the geometry, materials, and lighting of a 3D scene from images, also known as inverse rendering, has been a long-standing problem in the fields of computer vision and graphics.
It is gaining traction in this era of blowout VR and AR applications, where there is a high demand for easily acquired 3D contents from the real world. 
Previous capture systems, such as light-stages with controlled light directions and cameras\cite{lensch2003planned, guo2019relightables, zhang2021neural}, using a co-located flashlight and camera in a dark room\cite{Bi2020NeuralRF, Bi2020DeepRV}, and rotating objects with a turntable\cite{Dong2014Appearancefrommotion, Xia2016RecoveringSA}, show limitations in user-friendliness. 

More recent works\cite{physg2020, nerfactor, Boss2020NeRDNR} explore flexible capture settings under natural illumination. These methods typically represent geometry and spatially varying BRDF (SVBRDF) as coordinate-based neural networks and recover them by optimizing a re-rendering loss that compares rendered images with input images. 
However, capturing under natural illumination often shows complex effects such as soft shadows and interreflections. It is intractable to simulate these effects when optimizing SVBRDF and light parameters as it necessitates expensive recursive path tracing in physically based rendering. Prior methods usually ignore both self-occlusion and interreflection\cite{physg2020} in order to reduce computation, or only model visibility\cite{nerfactor} or limit the indirect lighting to a single bounce with known light sources\cite{Srinivasan2021NeRVNR}. 
Without properly modeling the indirect illumination, there exists a gap between the captured image and the rendered image. As a result, the effect of indirect illumination in the captured images is prone to being baked into the estimated diffuse albedo to compensate for this gap, as illustrated in Figure \ref{fig:banner}. 
It also results in artifacts in the recovered specular reflectance and environmental light as they explain the observed images together with albedo.

In this paper, we aim to estimate the SVBRDF of objects from multi-view RGB images captured under unknown static illumination. Our main technical innovation is an efficient approach to modeling indirect illumination in this inverse rendering process. We model the indirect illumination by a multilayer perceptron (MLP) that maps a 3D surface point to its indirect incoming illumination. The core idea to efficiently learning this indirect illumination MLP is that the indirect illumination doesn't need to be jointly learned with the SVBRDF and environmental light, but can be directly derived from the outgoing radiance field of the scene, which can be constructed from multi-view images with the off-the-shelf neural scene representation methods (e.g., \cite{Mildenhall2020NeRFRS,yariv2020idr}). 

Specifically, we first learn the geometry and outgoing radiance field of the object, both represented as MLPs, from the input images using the existing method \cite{yariv2020idr}. Then, the learned radiance field serves as the ground-truth incoming illumination of its reachable surface points to train the indirect illumination MLP.
Finally, the learned indirect illumination is plugged into the rendering equation and fixed during the optimization of SVBRDF and environmental light. In this way, the indirect illumination can be directly queried when optimizing the other unknowns without the need of recursive path tracing, making the inverse rendering problem better constrained and more efficient to solve. Furthermore, to reduce the ambiguity of disentangling BRDF and incident light, we introduce a prior that a real-world object should consist of limited types of materials. This prior is imposed by representing SVBRDF as an encoder-decoder with a sparse latent space.

We evaluate the proposed method on both synthetic and real datasets. The experimental results show that our approach outperforms baseline methods and is able to recover shadow- and interreflection-free albedo and high-quality roughness, as well as supporting realistic free-viewpoint relighting. 

%% file: 02_related_work.tex
\section{Background}

\paragraph{Inverse rendering.}
Inverse rendering, the task of decomposing the image appearance into the underlying intrinsic properties such as geometry, material, and lighting conditions, has been a longstanding problem in computer vision and graphics. 
The full inverse rendering problem in its most general form is well-known to be severely ill-posed.
The key problem in inverse rendering is to properly add priors and regularizations to the optimization process to mitigate the ill-posed condition.

Single-image inverse rendering methods~\cite{Barron2015ShapeIA,Li2020InverseRF,Li2018LearningTR,Lichy2021ShapeAM,Sang2020SingleShotNR,Sengupta2019NeuralIR,Wei2020ObjectbasedIE,Yu2019InverseRenderNetLS} rely heavily on the strong prior of the planar geometry. 
Since the planar input and output maps are naturally easier to be processed by CNNs, these methods can learn priors for normal, reflectance, and illumination from large-scale datasets. 
They can effectively infer plausible materials and normal maps from a single image but usually cannot recover spatially-varying 3D representations of these factors.

Most methods that recover fully factorized 3D geometry, materials and lighting require scenes to be captured under more constrained settings.
They either capture images while rotating the object with the camera fixed \cite{Xia2016RecoveringSA,Dong2014Appearancefrommotion}, or shoot a video using a handheld cellphone with a flash in a dark environment so that the point light is associated with the camera and its location is known\cite{Bi2020Deep3C,Bi2020DeepRV,Bi2020NeuralRF,Schmitt2020OnJE,Nam2018PracticalSA}.
The varied or known illumination provide rich information for inferring geometry and material properties.

\paragraph{Implicit neural representation.} 
Recent advances in implicit neural representation enable new possibilities for inverse rendering.
NeRF~\cite{Mildenhall2020NeRFRS} achieves photo-realistic novel view synthesis by representing radiance fields with multilayer perceptrons.
Supervised with differentiable volumetric rendering, NeRF is able to reconstruct the radiance field of a scene with only a collection of images. 
While NeRF represents geometry as volumetric density fields, some surface-based methods like IDR~\cite{yariv2020idr} and NeuS~\cite{wang2021neus} represent geometry with Signed Distance Functions (SDFs). 
These methods work well for novel view synthesis, but they only model the outgoing radiance of a surface and are not capable of disentangling it into the incoming radiance and the underlying material property. As a result, they don't enable free-viewpoint relighting.

\paragraph{Inverse rendering with implicit neural representation.} 
Enabled by the fully differentiable pipelines in implicit neural representation, recent methods in inverse rendering aim at more ``casual'' capture conditions.
Notably, PhySG~\cite{physg2020} and NeRFactor~\cite{nerfactor} decompose the scene under complex and unknown illumination. NeRD~\cite{Boss2020NeRDNR} extends NeRF to deal with captures under fixed or varying illumination.
However, all these methods only consider direct illumination from the light source and ignore indirect illumination, so they are unable to simulate interreflection effects to resemble the observed images. As a result, they can only model simple and convex surfaces that have neglectable indirect light. 
NeRV~\cite{Srinivasan2021NeRVNR} does consider one indirect bounce, but with known environment light and is rendered with Monte-Carlo ray tracing. Our method is able to reconstruct high quality indirect light with unconstrained bounces and does not require known lighting conditions.



\begin{figure*}[t]
\centering
\includegraphics[width=1\linewidth]{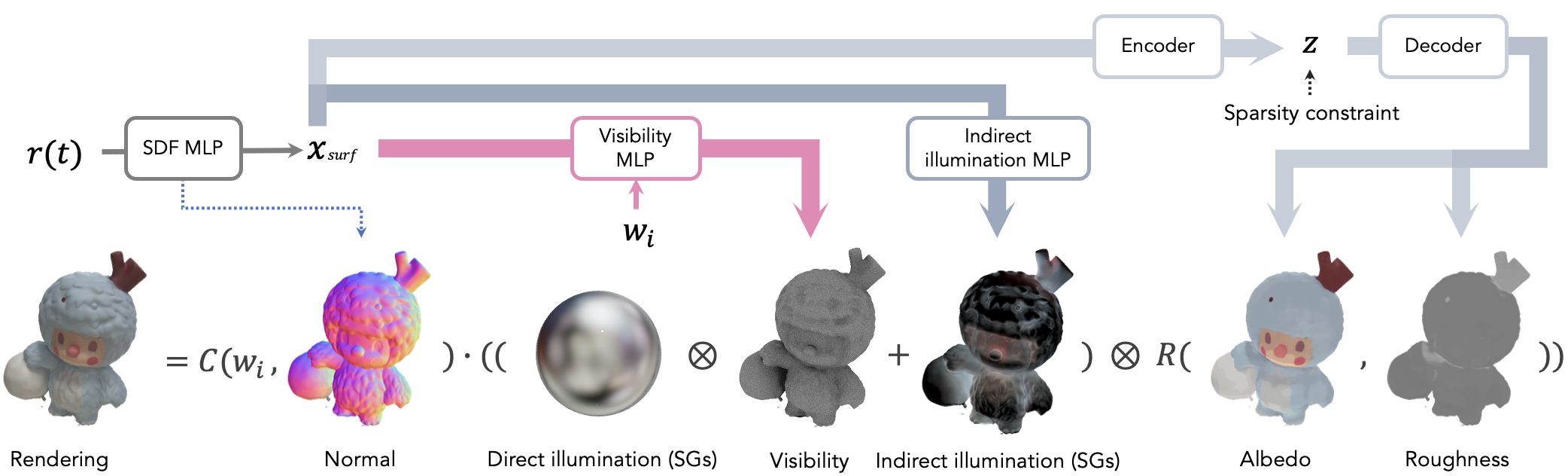}
\caption{\textbf{Forward rendering.} For a specific surface point $\hat{\mathbf{x}}$ , the full incoming light is modeled as the sum of direct illumination direction-wise multiplied by visibility and indirect illumination derived from the reconstructed outgoing radiance field. The spatially varying BRDF parameters are output from an encoder-decoder network with a sparsity constraint on the latent code, and each specular BRDF is further transformed to a single spherical Gaussian (SG). During forward rendering, only the BRDF and the direct illumination need to be optimized, while the others are all pre-acquired and fixed. In the bottom row, the visualized visibility is the mean value over all directions and the indirect illumination is the irradiance at each point.}
\label{fig:rendering}
\end{figure*}

\paragraph{The rendering equation.}

For non-emitted object, the rendering equation computes the outgoing radiance $L_o$ at surface point $\mathbf{\hat{x}}$ along direction $\boldsymbol{\omega}_o$ by integrating the reflected light over hemisphere\cite{kajiya1986rendering}:
\begin{equation}
\label{render_equ}
    L_o(\mathbf{\hat{x}}, \boldsymbol{\omega}_o) = \int_\Omega 
    L_{in}( \mathbf{\hat{x}}, \boldsymbol{\omega}_i)
    f_r(\mathbf{\hat{x}}, \boldsymbol{\omega}_i, \boldsymbol{\omega}_o)
    (\boldsymbol{\omega}_i \cdot \mathbf{n})\mathrm{d}\boldsymbol{\omega}_i
\end{equation}
The $L_{in}( \mathbf{\hat{x}}, \boldsymbol{\omega}_i)$ is the incoming radiance at surface point $\mathbf{\hat{x}}$ along direction $\boldsymbol{\omega}_i$ and the BRDF function $f_r$ describes how much light arriving from direction $\boldsymbol{\omega}_i$ is reflected towards direction $\boldsymbol{\omega}_o$ at $\mathbf{\hat{x}}$.

%% file: 03_method.tex
\section{Method}

\subsection{Overview}
Given a set of posed images of an object captured under static illumination, we learn to decompose the shape and SVBRDF to enable applications such as free-view relighting.
We solve the inverse rendering problem in an analysis-by-synthesis manner, where we optimize the parameters of the forward rendering model until the rendered images closely resemble the observed images.
Figure \ref{fig:rendering} depicts the forward rendering process of our proposed method.

In the paper, we represent the geometry as a zero level set as IDR \cite{yariv2020idr} by learning a Signal Distance Function (SDF), parameterized by a multilayer perceptron $S(\mathbf{x})$, that maps from a 3D location $\mathbf{x}$ to the SDF value at this location. It gives smooth and realistic surfaces of objects.
We decompose the spatially varying incoming light $L_{in}( \mathbf{\hat{x}}, \boldsymbol{\omega}_i)$ at a surface point $\mathbf{\hat{x}}$ along the direction $\boldsymbol{\omega}_i$ into two components: direct illumination $E$ evaluated by visibility (Sec. \ref{section:vis}) and indirect illumination $L_i$ efficiently derived from the outgoing radiance field (Sec. \ref{section:ind_illum}).
In contrast to previous works, the SVBRDF parameters in our formulation are parameterized as an encoder-decoder network with a sparse latent space (Sec. \ref{section:brdf}).

To render a camera ray, the intersection $\hat{\mathbf{x}}$ of the ray and SDF surface can be observed via the sphere tracing technique, and its corresponding surface normal is the gradient of the SDF: $\mathbf{n}=\nabla_\mathbf{\hat{x}} S$. Then we query visibility, indirect illumination, diffuse albedo and roughness from networks, and perform rendering together with environment lighting (Sec. \ref{section:render}). The parameters of SVBRDF and direct illumination are optimized by minimizing the reconstruction error between the renderings and the observed images.

\subsection{Visibility for Direct Illumination}
\label{section:vis}
For direct illumination, we assume that all lights come from an infinitely faraway environment and parameterize them as M=128 spherical Gaussians (SGs)\cite{wang2009all}:
\begin{equation}
    E(\boldsymbol{\omega_i}) =\sum_{k=1}^{M} G(\boldsymbol{\omega_i}; \boldsymbol{\xi}_k, \lambda_k, \boldsymbol{\mu}_k)
\end{equation}
 where $\boldsymbol{\xi}\in \mathbb{S}^2$ is the lobe axis, $\lambda\in  \mathbb{R}_+$ is the lobe sharpness, and $\boldsymbol{\mu} \in \mathbb{R}^3$ is the lobe amplitude. 

The environment lighting is evaluated by the visibility indicating whether the direction $ \boldsymbol{\omega}_i$ at surface point $\mathbf{x}$ is occluded or not. The visibility can be obtained by performing sphere tracing from surface points to light sources. However, the tracing step is repeatedly executed during forward rendering and is time-consuming. So we re-parameterize it as an MLP that maps the surface point location $\mathbf{x}$ and direction $\boldsymbol{\omega}_i$ to visibility: $V (\mathbf{x}, \boldsymbol{\omega}_i) \mapsto v$. The network provides a compact and continuous representation and requires only a small number of sampled rays above surface points for training.
The direction-wise multiplication of the visibility function and the environment lighting SG is desired to yield another SG to support the integral of the spherical functions during rendering. We achieve this by having the amplitude of the output SG produce the same integrated value as the original lobe and preserving its center:
\begin{equation}
     V(\mathbf{x}, \boldsymbol{\omega}_i) \otimes G(\boldsymbol{\omega_i}; \boldsymbol{\xi},
    \boldsymbol{\lambda}, \boldsymbol{\mu})
     \approx G(\boldsymbol{\omega_i}; \boldsymbol{\xi},
    \boldsymbol{\lambda},  \gamma\boldsymbol{\mu})
\end{equation}
\begin{equation}
    \gamma = \frac{\sum^S_{k=1} G(\boldsymbol{\omega}_k)V(\mathbf{x}, \boldsymbol{\omega}_k)}{\sum^S_{k=1} G(\boldsymbol{\omega}_k)}
\end{equation}
The visibility ratio $\gamma$ is obtained by randomly sampling the $S=32$ directions in the SG lobe and taken a weighted average of queried visibility.

\begin{figure}[t]
\centering
\includegraphics[width=1\linewidth]{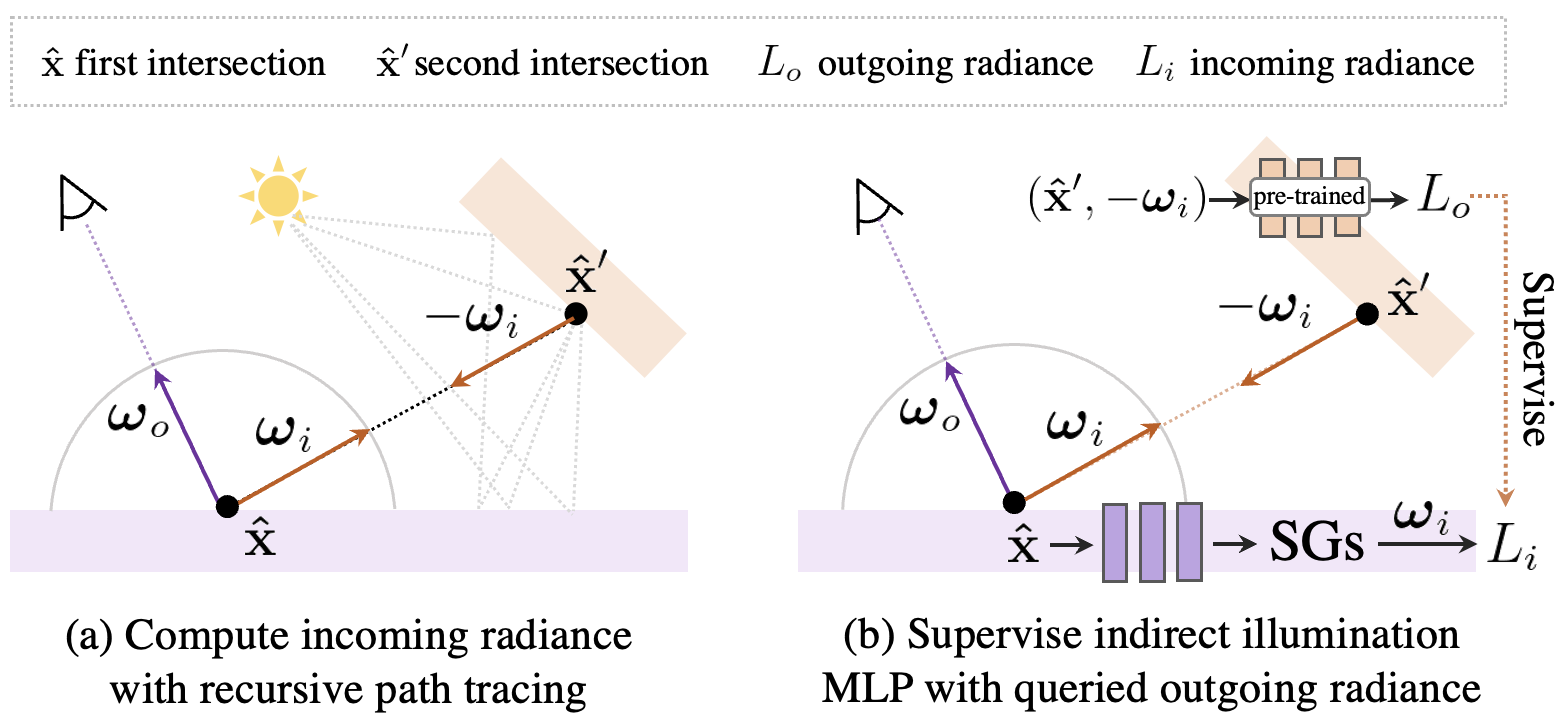}
\caption{Instead of computing incoming radiance by performing costly recursive path tracing (a), we consider the pre-trained outgoing radiance field as indirect illumination and train a network that maps a 3D location to its indirect incoming illumination represented as a mixture of SGs (b).}
\label{fig:lighting}
\end{figure}

\subsection{Indirect Illumination}
\label{section:ind_illum}
According to the rendering equation, the indirect incoming radiance $L_i( \mathbf{\hat{x}}, \boldsymbol{\omega}_i)$ at the intersection $\mathbf{\hat{x}}$ of the camera ray and surface toward direction $\boldsymbol{\omega}_i$ is obtained by first performing ray tracing, and then assigned by the outgoing radiance $L_o(\mathbf{\hat{x}'}, -\boldsymbol{\omega}_i)$ of the second intersection $\mathbf{\hat{x}'}$ toward direction $-\boldsymbol{\omega}_i$:
\begin{equation}
    L_i( \mathbf{\hat{x}}, \boldsymbol{\omega}_i) = L_o(\mathbf{\hat{x}'}, -\boldsymbol{\omega}_i)
\end{equation}
$L_o(\mathbf{\hat{x}'}, -\boldsymbol{\omega}_i)$ is rendered by continuing sampling and integrating rays over the hemisphere, as illustrated in Figure \ref{fig:lighting}. As the number of considered bounces increases, the tracing and rendering computation grows with the  \textit{exponential} order of the sample amount.
It is typically intractable in reality and increases the complexity of decomposing unknowns from rendering. 

We tackle this problem by reconstructing the outgoing radiance field and deriving indirect illumination from it, rather than performing exhaustive ray tracing for the indirect illumination. The outgoing radiance field, which can be viewed as a neural renderer, is a continuous function of the surface point location $\mathbf{\hat{x}}$, normal $\mathbf{\hat{n}}$ and viewing direction $\boldsymbol{\omega_o}$: $R(\mathbf{\hat{x}}, \mathbf{\hat{n}}, \boldsymbol{\omega_o})  \mapsto L_o$. We learn this field parameterized as an MLP from observed images together with geometry using view synthesis method\cite{yariv2020idr}. 
Therefore, the outgoing radiance of the second intersection, which is the cumulative results of multiple bounces, is obtained by querying the MLP:
\begin{equation}
     L_o(\mathbf{\hat{x}'}, -\boldsymbol{\omega}_i) = R(\mathbf{\hat{x}'}, \mathbf{\hat{n}'}, -\boldsymbol{\omega}_i)
\end{equation}
where $\mathbf{\hat{n}'}$ is the normal of the second intersection.

 We further transfer it into indirect illumination represented as a mixture of SGs and cache it in an MLP to avoid duplicate computation of tracing from $\mathbf{\hat{x}}$ to $\mathbf{\hat{x}'}$. The representation facilitates the hemispherical integration with other SG lobes, thus avoiding the use of the Monte-Carlo method, which requires a trade-off between low-cost sampling and high-quality rendering.
 Here, we introduce the indirect illumination MLP $I(\mathbf{x})$ that outputs the SG parameters $\Gamma \in  \mathbb{R}^{24\times7}$ at any input 3D location $\mathbf{x}$. 
 The incoming radiance is determined by querying the SG function at the desired surface point and direction:
\begin{equation}
     L_i(\mathbf{\hat{x}}, \boldsymbol{\omega}_i) = G(\boldsymbol{\omega}_i; I(\mathbf{\hat{x}}))
\end{equation}
The indirect illumination MLP is supervised by first drawing samples from outgoing radiance field $R$, and then forcing the incoming radiance to reproduce the corresponding outgoing radiance. We visualize this process in Figure \ref{fig:lighting}.






\subsection{BRDF}
\label{section:brdf}
 We use the simplified Disney BRDF model\cite{burley2012physically} with diffuse albedo $\mathbf{a}$ and roughness $\mathbf{r}$ as parameters and assume dielectric materials with fixed $F_0=0.02$ in the Fresnel term.
 
Parameterizing SVBRDF by directly mapping surface points to its parameters is straightforward. However, it often leads to noisy roughness since a few surface points lack supervision due to the distribution of the training views or self-occlusion. We alleviate this problem by introducing a prior that an object is usually composed of a small amount of materials.

Our solution is to represent SVBRDF as an encoder-decoder network with a sparse latent space.
The network transforms the input surface point $\mathbf{x}$ to its corresponding latent code $\mathbf{z}$ and decodes it to its diffuse albedo and roughness.
We impose a sparsity constraint\cite{ng2011sparse} on the latent code so that most of the channels in $\mathbf{z}$ are close to zero:
\begin{equation}
    \ell_{\rm KL} = \sum_{j=1}^{n}{\rm KL}(\rho\parallel \hat{\rho_j})
\end{equation}
where ${\rm KL}(\rho\parallel \hat{\rho_j}) = \rho{\rm log}\frac{\rho}{\hat{\rho_j}} + (1-\rho){\rm log}\frac{1-\rho}{1-\hat{\rho_j}}$ is a Kullback-Leibler divergence loss and $\hat{\rho_j}$ is the average of $j^{th}$ channel of $\mathbf{z}$ over batch input. $\rho$ is set to 0.05. $n$ is the length of latent code. We further apply a smooth loss on the decoder $D$ such that close latent codes are clustered to yield same SVBRDF:

\begin{equation}
    \ell_{\rm s} = || D(\mathbf{z}) - D(\mathbf{z} + \boldsymbol{\xi}) ||_1
\end{equation}
where $\boldsymbol{\xi}$ is a small random variable drawn from a normal distribution with zero mean and 0.01 variance.

\subsection{Rendering}
\label{section:render}
The BRDF function $f_r$ in Equation \ref{render_equ} contains a diffuse component $\frac{\mathbf{a}}{\pi}$ and a specular component $f_s(\mathbf{\hat{x}}, \boldsymbol{\omega}_i, \boldsymbol{\omega}_o)$.
We convert both the specular BRDF $f_s$ and the clamped cosine factor $C=\boldsymbol{\omega}_i \cdot \mathbf{n}$ to a single SG as in prior work \cite{physg2020}. So Equation \ref{render_equ} can be approximated as the fast inner product of SGs. Specifically, we separate the rendering of direct illumination into diffuse component $L_d$ and specular component $L_s$. The diffuse component is calculated as the sum of the integrals of each masked environment lighting SG and the clamped cosine factor:
\begin{equation}
    L_d(\hat{\mathbf{x}}) = \frac{\mathbf{a}}{\pi} \sum_{k=1}^M
    (V(\hat{\mathbf{x}}, \boldsymbol{\omega}_i) \otimes E_k(\boldsymbol{\omega}_i))  \cdot C\\
\end{equation}
Note that in the specular component, in order to accurately approximate the final integral in the presence of a narrow specular lobe, the visibility ratio $\gamma$ is determined by sampling the specular SG:
\begin{equation}
    L_s(\hat{\mathbf{x}}, \boldsymbol{\omega}_o) = \sum_{k=1}^M
    (f_s \otimes V(\hat{\mathbf{x}}, \boldsymbol{\omega}_i)) \otimes E_k(\boldsymbol{\omega}_i) \cdot C
\end{equation}

As for the rendering of indirect illumination, the spatially varying indirect illumination is first queried from the indirect illumination MLP $I$, and the rendering is similar to the above process, except that the visibility is not required.

\subsection{Training}
We optimize the geometry, SVBRDF and environment lighting from a set of posed images through three-stage training. First, the SDF MLP $S(\mathbf{x})$ and outgoing radiance MLP $R$ are optimized using \cite{physg2020}. Second, we sample 256 surface points and draw 16 sampled rays for each, then perform sphere tracing to obtain visibility and incoming radiance simultaneously, which serve as the ground truth for supervising the visibility MLP $V$  and indirect illumination MLP $I$
via cross-entropy loss and $\ell_1$ loss. Last, the diffuse albedo, roughness and direct illumination are jointly optimized by minimizing the reconstruction loss $\ell_{\rm recon}$ between the renderings and the observed images.
The full loss in the final stage is:
\begin{equation}
    \ell =  \lambda_{\rm recon}\ell_{\rm recon} + \lambda_{\rm KL}\ell_{\rm KL} + \lambda_{\rm s}\ell_{\rm s}
\end{equation}
We set weights $\lambda_{\rm recon}=1.0$, $\lambda_{\rm KL}=0.01$, $\lambda_{\rm s}=0.1$ in our experiments.

The architecture of visibility MLP, indirect illumination MLP and encoder of BRDF contains 4 layers with 512 hidden units. Positional encoding\cite{Bi2020NeuralRF} is applied to the input 3D locations and directions with 10 and 4 frequency components, respectively. The decoder of BRDF is a 2-layer network with a 32-dimensional input latent code and 128 hidden units. We implement our model in PyTorch and optimize using Adam\cite{kingma2014adam} with learning rate $5e^{-4}$. Both of the latter two stages run 200 epochs on a single RTX 3090 GPU, which takes about 1 and 2 hours, respectively.


%% file: 04_experiment.tex
\begin{figure*}[t]
\centering
\includegraphics[width=1\linewidth]{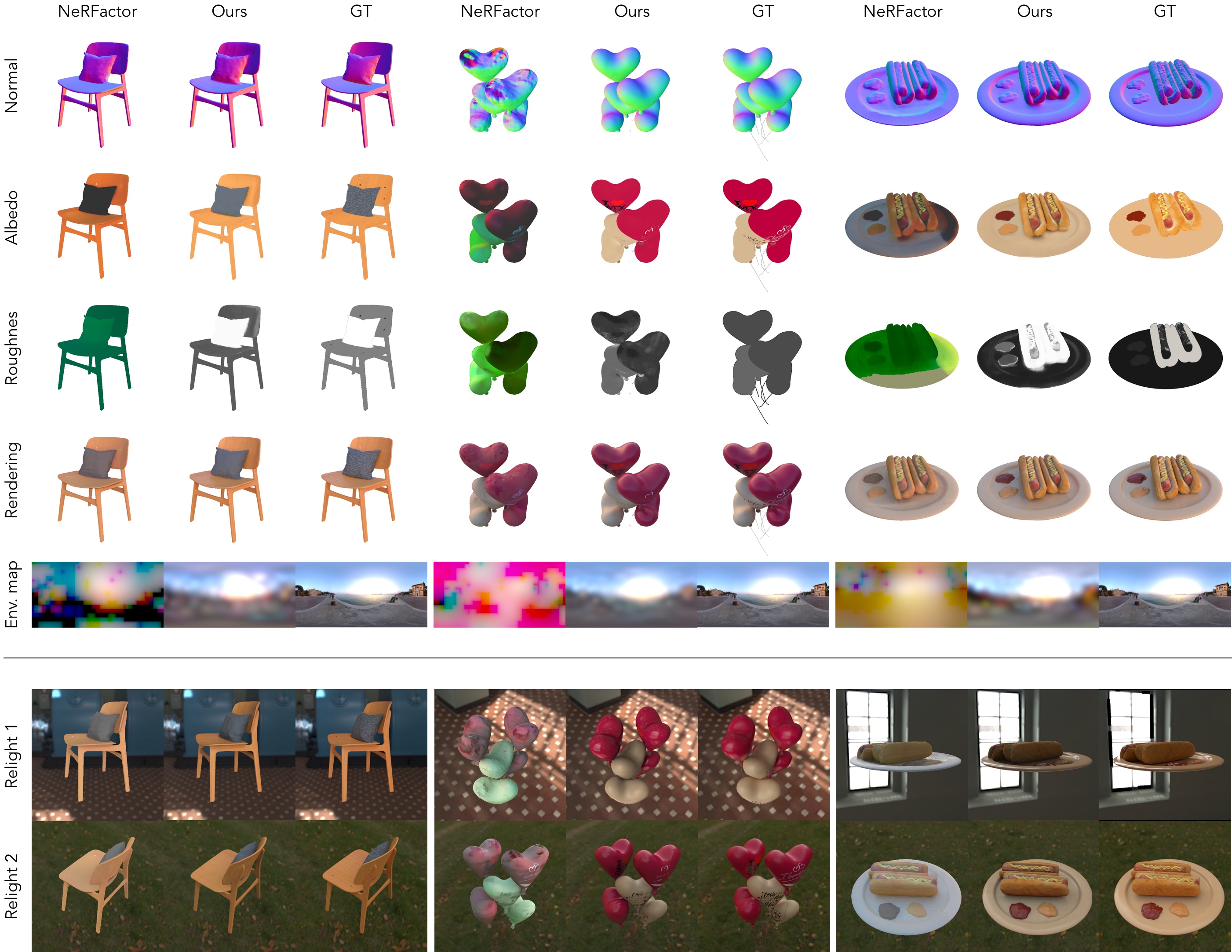}
\caption{\textbf{Comparisons with previous work.} We visualize the estimated normal, diffuse albedo, roughness and environment map of NeRFactor\cite{nerfactor} and our method on three scenes. Note that the roughness of NeRFactor is visualized with the latent code, which represents a BRDF identity, since it is parameterized by a learned model. We also compare the re-renderings under a novel view and original light (the fourth row) as well as novel views and novel light (the last two rows).
}
\label{fig:comparison}
\end{figure*}

\section{Experiments}
 In this section, we conduct experiments to investigate the performance of our inverse rendering approach. First, we briefly present how we build a synthetic dataset to examine our setting in Sec.~\ref{subsec:dataset}. Then, we make quantitative comparisons with two baselines on the synthetic data in Sec.~\ref{subsec:comparison}. Third, we perform several ablations to discuss our key components in Sec.~\ref{subsec:ablation}. Finally, we qualitatively study the inverse rendering and relighting abilities of our method on the real dataset in Sec.~\ref{subsec:real}. We refer to the supplemental materials for more results. 

\subsection{Synthetic Data}
\label{subsec:dataset}
We collect 4 CAD models, each with obvious self-occlusions and multiple materials.
For a specific object, we assign it with a natural environment map, and render 100 training images as well as their masks via Blender Cycles. Masks are required by the SDF learning process \cite{yariv2020idr}. 
We render other 200 test images as well as their albedo and roughness maps to evaluate the novel view synthesis performance and the inverse rendering ability. 
To measure the relighting performance, we utilize other two environment maps and render 200 images for each case. The image resolution is $800\times800$.


\subsection{Baseline Comparisons}
\label{subsec:comparison}
To our best knowledge, there are only a few works that study the exactly same inverse rendering setting as this paper, \emph{i.e,} training with fixed unknown illumination while supporting free-view relighting. We take NeRFactor \cite{nerfactor} and PhySG \cite{physg2020} as baselines and make quantitative comparisons on the synthetic datasets. The image quality metrics include Peak Signal-to-Noise Ratio (PSNR), Structural Similarity Index Measure (SSIM), and Learned Perceptual Image Patch Similarity (LPIPS)\cite{zhang2018unreasonable}. Since there is an inevitable scale ambiguity in estimating the albedo and environment lighting, we additionally evaluate the albedo after aligning with the ground truth, as done in \cite{physg2020,nerfactor}.

NeRFactor distills the volumetric geometry of NeRF \cite{Mildenhall2020NeRFRS} into a surface representation. It relies on a data-driven BRDF prior learned from real-world BRDF measurements to recover 3D neural fields of SVBRDF. Table \ref{tab:bc} and Figure \ref{fig:comparison} demonstrate that our method is superior to NeRFactor both quantitatively and qualitatively. NeRFactor parameterizes illumination as a $16 \times 32$ resolution environment map so that each pixel/parameter can vary independently. The estimated results show that the albedo would easily be baked into the environment map during its optimization process, thus resulting in poor relighting performance. In contrast, our predicted environment lighting and SVBRDF contain fine details and are visually close to the ground truth maps, as shown in Figure~\ref{fig:comparison}. 

PhySG is able to jointly recover environmental lighting, BRDFs and geometry from multi-view inputs captured under static illumination. However, it presumes that the recovered object is homogeneous.
We adapt its pipeline by replacing its global roughness with a spatially varying roughness parameterized as an MLP. The experimental results in Table \ref{tab:bc} show that it performs badly. The main reason is that, without modeling visibility and indirect illumination, geometry optimization is highly ill-posed, especially in areas with obvious shadow and interreflection.


\begin{table*}[t]
\small
\begin{center}

\resizebox{\textwidth}{16mm}{
\begin{tabular}{ c c c c c c c c c c c c c c c c c}
\toprule
& Roughness 
& \multicolumn{3}{c}{Albedo} & \multicolumn{3}{c}{Aligned Albedo} &
 \multicolumn{3}{c}{View Synthsis} & \multicolumn{3}{c}{Relighting}\\
 \cmidrule(lr){2-2}\cmidrule(lr){3-5}\cmidrule(lr){6-8}\cmidrule(lr){9-11} \cmidrule(lr){12-14}
 Method & MSE $\downarrow$ 
 & PSNR $\uparrow$ & SSIM $\uparrow$ & LPIPS $\downarrow$ 
 & PSNR $\uparrow$ & SSIM $\uparrow$ & LPIPS $\downarrow$ 
 & PSNR $\uparrow$ & SSIM $\uparrow$ & LPIPS $\downarrow$
 & PSNR $\uparrow$ & SSIM $\uparrow$ & LPIPS $\downarrow$ \\
\midrule
NeRFactor\cite{nerfactor} & - &
19.4858 & 0.8641 & 0.2060 &
22.9647 & 0.9064 & 0.1617 & 
22.7953 & 0.9168 & 0.1512 & 
21.5373 & 0.8749 & 0.1708 \\ 

PhySG*\cite{physg2020} & 0.2682 &
21.2690&0.9722&0.0962&
21.7968&0.9733&0.1845&
23.4154&0.9871&0.0684&
22.6288&0.9734&0.0726 \\ 
\midrule

Ours & \textbf{0.0723} &
\textbf{24.1608} & \textbf{0.9782} & \textbf{0.0566} & 
\textbf{25.2511} & \textbf{0.9825} & \textbf{0.0581} & 
26.1918 & 0.9905 & 0.0438 & 
\textbf{25.5934} & \textbf{0.9840} & \textbf{0.0410} \\

w/o vis. \& ind. illum. & 0.1575 &
23.3332 & 0.9758 & 0.0674 & 
24.0401 & 0.9720 & 0.0679 & 
\textbf{26.4971} & 0.9923 & 0.0437 & 
25.3919 & 0.9804 & 0.0451 \\

w/o ind. illum. & 0.0845 &
23.7422 & 0.9731 & 0.0677 &
24.6547 & 0.9819 & 0.0651 & 
26.3454 & \textbf{0.9927} & \textbf{0.0435} & 
25.4957 & 0.9836 & 0.0444 \\

w/o latent space & 0.0783 &
24.0930 & 0.9775 & 0.0593 &
25.2283 & 0.9824 & 0.0598 & 
26.1846 & 0.9902 & 0.0449 & 
25.5101 & 0.9837 & 0.0422 \\
\bottomrule
\end{tabular}}
\end{center}
\vspace{-4mm}
\caption{\textbf{Quantitative evaluations.} We present the average results on the test images of all four synthetic scenes. ``Aligned Albedo" refers to scaling the albedo prediction for each RGB channel to match the ground truth before computing the errors. We slightly modify PhySG\cite{physg2020} to adapt it to our data by outputting spatially varying roughness from an MLP instead of treating it as a global variable.
Compared with previous methods and baseline models, our full model achieves the best performance in SVBRDF recovery and relighting. The view synthesis quality of the full model is slightly worse than the baselines, likely due to the rendering noise introduced by the visibility sampling.}
\label{tab:bc}
\end{table*}

\begin{figure*}[t]
\centering
\includegraphics[width=1\linewidth]{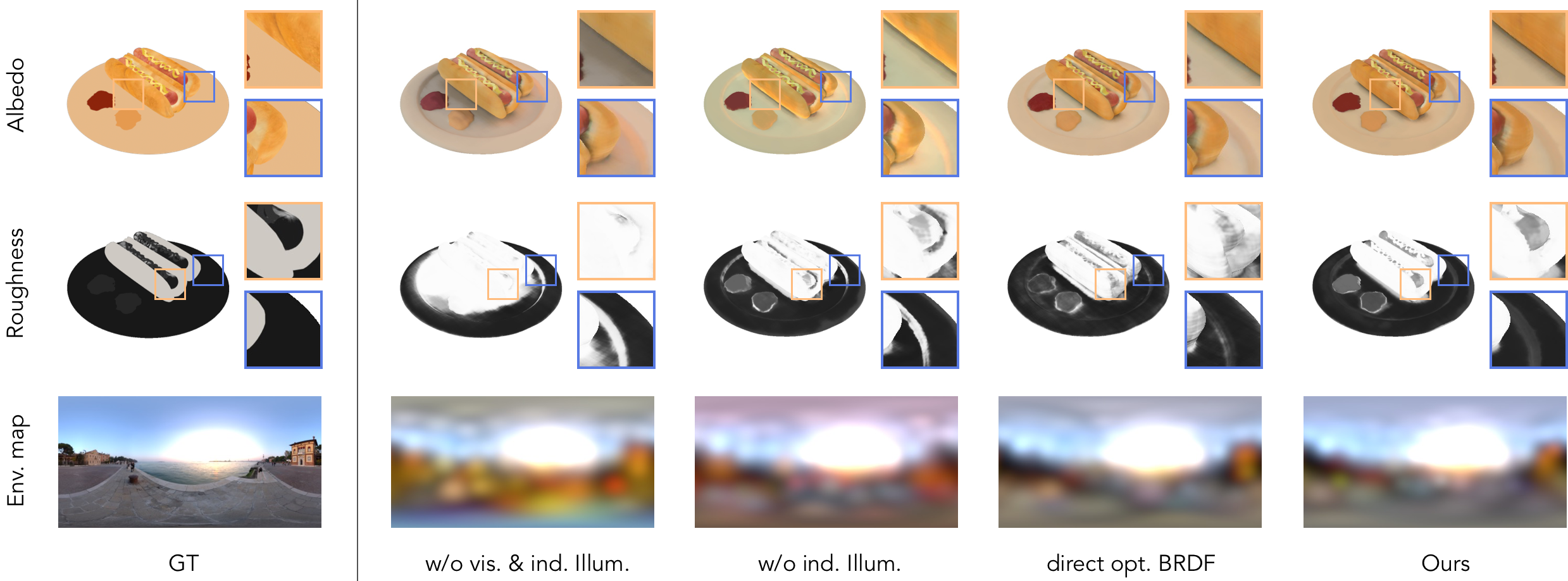}
\caption{\textbf{Ablation study on a synthetic scene (hotdog).} Please refer to Section \ref{subsec:ablation} for detailed descriptions.
}
\label{fig:ablation}
\end{figure*}

\begin{figure*}[t]
\centering
\includegraphics[width=1\linewidth]{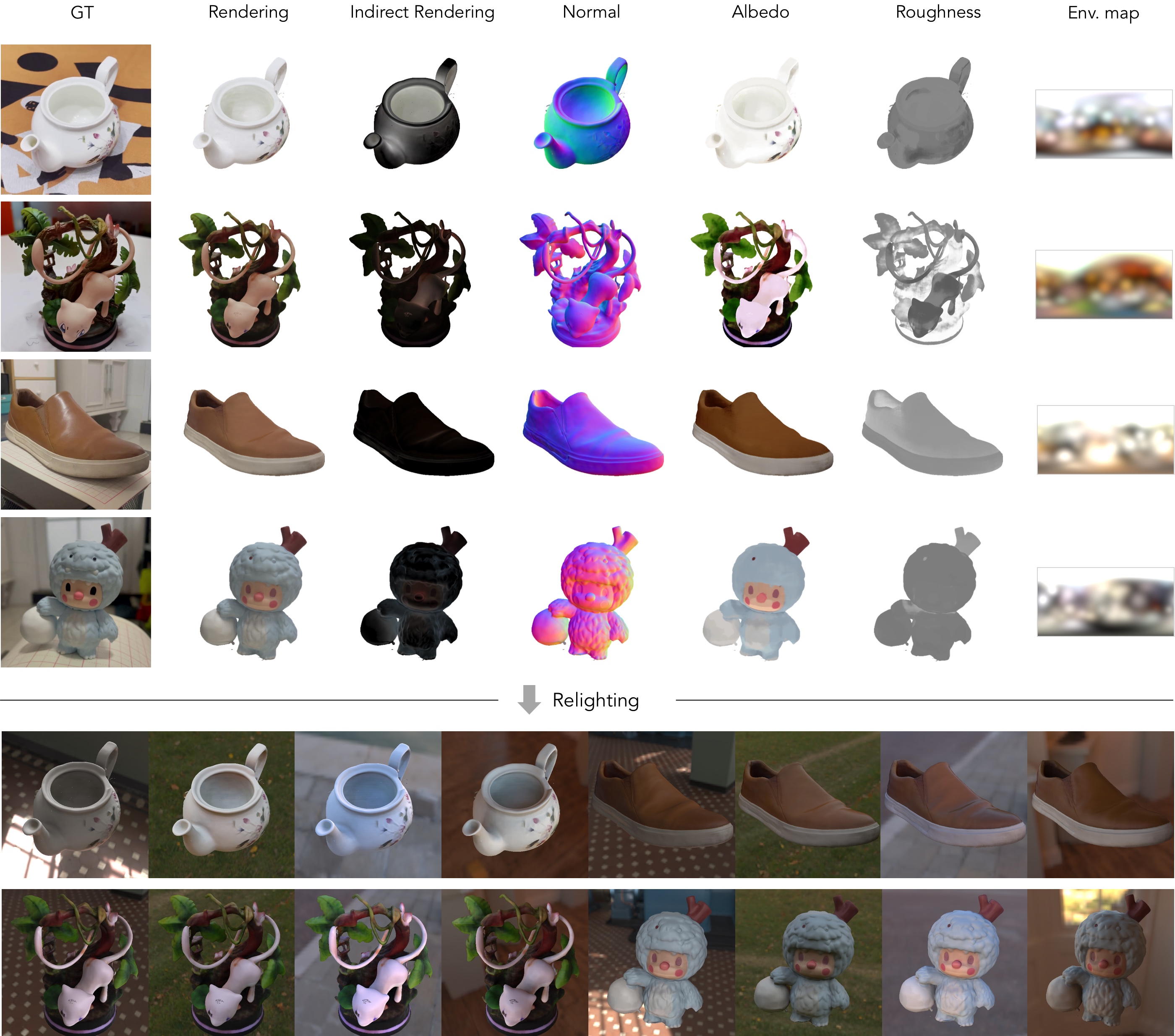}
\caption{\textbf{Results on real captures.}  Our method is capable of dealing with real-world objects composed of multiple materials. For each captured object, we show an image in the test set, 
our rendering, rendering under indirect illumination with our estimated shape and SVBRDF,  decomposed normal, albedo and roughness.
With decomposed factors, we can relight the object under arbitrary lighting. Here we show the results under four novel real-world illuminations.
}
\label{fig:real_cap}
\end{figure*}

\subsection{Ablation Studies}
\label{subsec:ablation}
We ablate combinations of three components of our methods that primarily affect the inverse rendering quality. We argue that a slight improvement over the studied metrics may bring an upgraded visual experience as rendering is a detailed effect. The results are reported in Table~\ref{tab:bc} and Figure~\ref{fig:ablation}.

In ``w/o vis. \& ind. illum.", we train a model under the assumption that all the surface points share the same environment lighting. It does not involve indirect illumination and visibility factors. It's not surprising that this variant performs worst for inverse rendering and relighting tasks. However, it yields a slight improvement in novel view synthesis. A possible reason is that visibility sampling introduces some rendering noise.
``w/o ind. illum." produces unexpected brighter environment lighting and albedo compared to ground truth. That means, without modeling the indirect illumination, these indirect lighting effects would be baked into the estimated albedo by mistake. The ``w/o latent space" variant trains an MLP that directly maps a 3D location to its diffuse and roughness using a re-rendering loss only, without latent space assumption (See Sec~\ref{section:brdf}).
The visualization shows that optimizing each surface point independently yields noise roughness.





\subsection{Results on Real Captures.} 
\label{subsec:real}
We select 4 real objects made of various materials, such as plastic and leather, and capture them with a mobile phone moving around the upper hemisphere. The camera poses are estimated by COLMAP\cite{Schonberger_2016_CVPR}. For each object, we uniformly sample 100 frames from the video and apply inverse gamma correction ($\gamma=2.2$) to the images for training. Note that the environment may not be exactly ideal, as not all light is infinite distance, especially when capturing object-centric video indoors, and moving people will cast shadows on objects. Figure \ref{fig:real_cap} shows the inverse rendering and relighting results. Our approach is able to infer plausible SVBRDF and support realistic relighting. See supplementary video for more results.

%% file: 05_conclusion.tex
\section{Conclusion}

In this paper, we present a novel approach to efficiently modeling the indirect illumination in the inverse rendering task. Most of the previous methods have not considered indirect illumination since simulating it is intractable within a physically-based rendering framework. Instead, we utilize the neural outgoing radiance field and derive indirect illumination from it. We demonstrate that, together with our proposed BRDF prior and SG-based visibility estimation, the full pipeline is able to estimate high-quality albedo and roughness from multi-view images captured under natural illumination and support realistic relighting.

\vspace{-0.5em}

\paragraph{Limitations.}
Our approach has the following limitations. First, our pipeline strongly relies on fine geometry as an input. We cannot deal with the case where the geometry fails to be reconstructed using \cite{yariv2020idr}. Fortunately, our rendering model can be easily migrated to other surface-based geometric representations.
Second, we parameterize BRDF with fixed $F_0=0.02$ in the Fresnel term\cite{burley2012physically}. In other words, we assume that the recovered materials are dielectric. Making $F_0$ learnable would exacerbate the ambiguity of the inverse problem. Learning-based prior or extra observations can help alleviate this ambiguity. We leave this as future work.